%% file: main.tex
\pdfoutput=1

\documentclass[11pt]{article}

\usepackage{acl}

\usepackage{times}
\usepackage{latexsym}
\usepackage{tabularx}
\usepackage{booktabs}
\usepackage{graphicx}
\usepackage{paralist}
\usepackage{multicol}
\usepackage{multirow,makecell}

\newcommand{\smallcitep}[1]{\footnotesize\citep{#1}}
\renewcommand{\paragraph}[1]{\vspace{1.25mm}\noindent\textbf{#1}}

\usepackage[T1]{fontenc}

\usepackage[utf8]{inputenc}

\usepackage{microtype}
\usepackage{amsfonts,amsmath,bm,xspace}

\newcommand{\x}{\bm x}
\newcommand{\y}{\bm y}
\newcommand{\hit}{\checkmark}

\title{Helping the Weak Makes You Strong: Simple Multi-Task Learning Improves Non-Autoregressive Translators}

\author{Xinyou Wang$^{\spadesuit}$\quad Zaixiang Zheng$^{\heartsuit}$\quad Shujian Huang$^{\spadesuit}$\\
$^{\spadesuit}$National Key Laboratory for Novel Software Technology, Nanjing University \\
$^{\heartsuit}$ByteDance AI Lab \\
{\textit{wangxinyou@smail.nju.edu.cn}, \textit{zhengzaixiang@bytedance.com}} \\
{\textit{huangsj@nju.edu.cn}}
}

\begin{document}
\maketitle

\input{00abstract}
\input{01introduction}
\input{02preliminary}
\input{03methodology}
\input{04experiments}
\input{05conclusion}
\input{06limitation}

\section*{Acknowledgements}

We would like to thank the anonymous reviewers for their insightful comments.
Meanwhile, we also want to thank Yu Bao for his valuable suggestions.
Zaixiang Zheng and Shujian Huang are corresponding authors.
This work is supported by National Science Foundation of China (No. U1836221, 6217020152).

\bibliography{anthology,custom}
\bibliographystyle{acl_natbib}

\input{07appendix}

\end{document}

%% file: 00abstract.tex
\begin{abstract}
    Recently, non-autoregressive (NAR) neural machine translation models have received increasing attention due to their efficient parallel decoding.
    However, the probabilistic framework of NAR models necessitates conditional independence assumption on target sequences, falling short of characterizing human language data.
    This drawback results in less informative learning signals for NAR models under conventional MLE training, thereby yielding unsatisfactory accuracy  compared to their autoregressive (AR) counterparts.
    In this paper, we propose a simple and model-agnostic multi-task learning framework to provide more informative learning signals.
    During training stage, we introduce a set of sufficiently weak AR decoders that solely rely on the information provided by NAR decoder to make prediction, forcing the NAR decoder to become stronger or else it will be unable to support its weak AR partners.
    Experiments on WMT and IWSLT datasets show that our approach can consistently improve accuracy of multiple NAR baselines without adding any additional decoding overhead.
    \footnotetext[1]{Code will be released at \url{https://github.com/wxy-nlp/MultiTaskNAT}.}
\end{abstract}

%% file: 01introduction.tex
\section{Introduction}

State-of-the-art neural machine translation (NMT) systems are mainly autoregressive~(AR) models~\cite{bahdanau2014neural,vaswani2017attention}, which decompose the joint probability of a sequence of tokens in a left-to-right order, modeling dependencies of each token with its preceding ones.
Despite having strong performance, such sequential decoding causes considerable latency, thereby unsatisfactory efficiency.

\begin{figure}[t]
  \centering
  \small
  \resizebox{1\linewidth}{!}{%
  \includegraphics[width=0.95\textwidth]{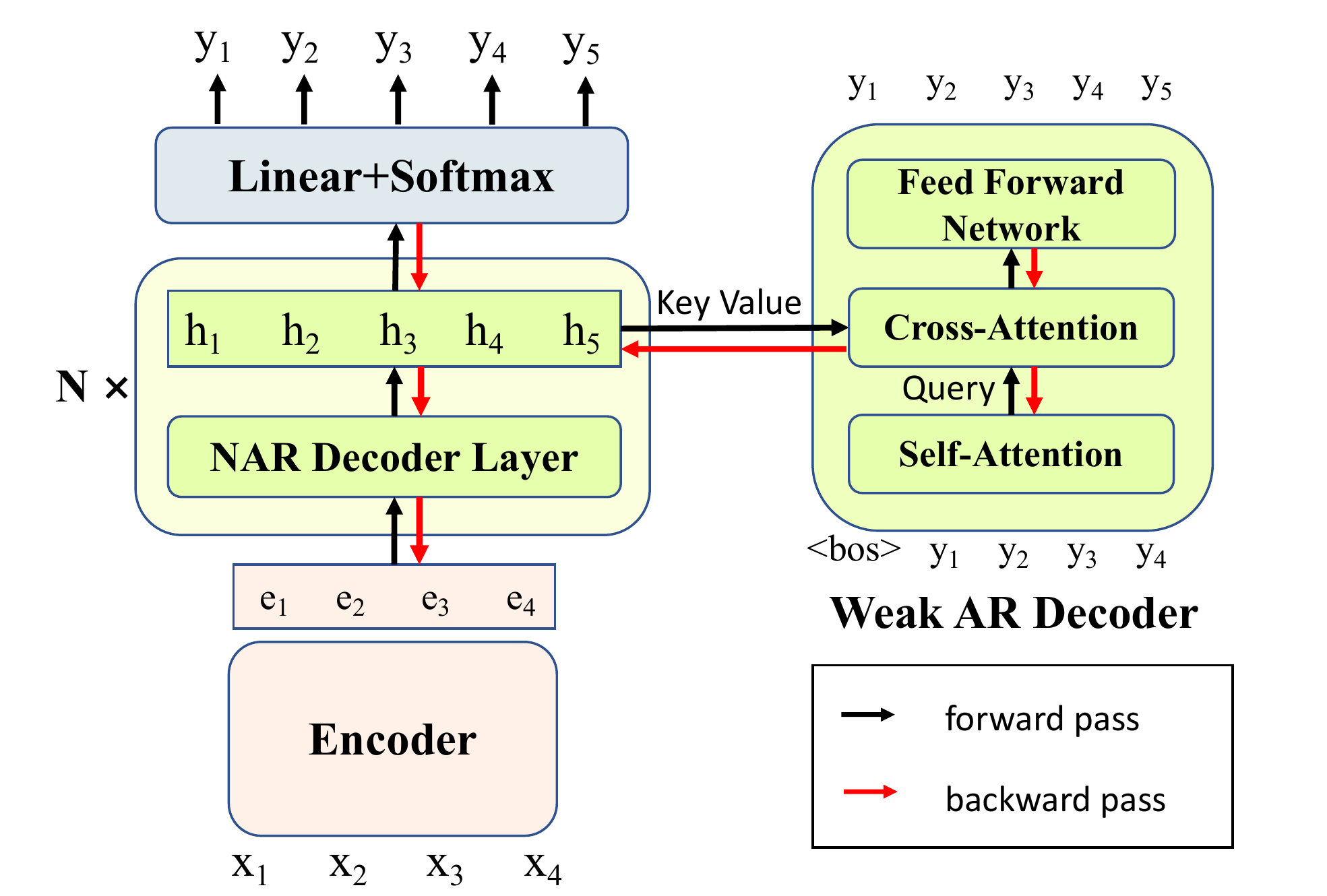} 
  }
  \caption{Illustration of our approach, where we introduce a set of auxiliary weak AR decoders, each of which must make its predictions solely relying on the information contained in the NAR decoder hidden states.
  Thus, the information provided by the NAR decoder must be sufficiently useful for the AR decoders to be capable of predicting the target sequence because the AR decoders are parameterized as weakly as possible, which will in turn let the NAR decoder learn to get stronger.}
  \label{mtl-multi}
  \vspace{-17pt}
\end{figure}

In contrast, non-autoregressive (NAR) translation models~\cite{gu2017non} permit potentially more efficient parallel decoding.
To do so, NAR models necessitate a notorious conditional independence assumption on target sequences as a trade-off.
This assumption, however, is probalistically insufficient to describe the highly multi-modal nature of human language data, imposing severe challenges for NAR models in a way of yielding less informative learning signals and gradients under the conventional MLE training.
As a result, NAR models often manifest implausible neural representations, especially in the decoder part as the decoder governs the generation, resulting in significant performance sacrifice. 
To close the accuracy gap, a majority of previous studies aim at improving the modeling of dependencies with more conditional information~\cite{qian2020glancing, ghazvininejad2019mask}. 
We argue that these research efforts are equivalent to providing better alternative learning signals without changing the NAR models' probabilistic framework.
However, most of these methods require a specific modification to the commonly-used Transformer model architecture.


A natural question may arise: can we encourage the NAR decoder to learn from sources of signals that are more informative than that of the conditional independence assumption, in order to better capture target dependencies? It would be more advantageous if it is also modification-free regarding model architectures and could also used with all current NAR systems.

In this paper, we propose a simple multi-task learning framework that introduces auxiliary weak AR decoders to make NAR models stronger.
The key idea is that we parameterize the auxiliary AR decoders as weakly as possible and force them to predict target sequences solely based on the information from NAR decoder's hidden representations, such that they can no longer model the target sequence on their own unless the knowledge provided by the NAR decoder is sufficiently useful.
As a result, the NAR decoder has no choice but to become stronger so as to support the AR partners that are poorly parameterized.
Additionally, our approach is plug-and-play and model-agnostic, and the weak AR decoders that we introduce are discarded during the inference stage, resulting in no additional decoding overhead.

We empirically evaluate its applications to several classes of NAR model, including vanilla NAR Transformer~\cite{gu2017non} and its CTC-based variant~\cite{libovicky2018end,saharia2020non}.
Experiments on widely-used WMT14 English-to-German, WMT16 English-to-Romanian, and IWSLT14 German-to-English benchmarks show that our approach consistently helps build more accurate NAR models over strong baselines.

%% file: 02preliminary.tex
\section{Preliminary}

Neural machine translation~(NMT) is formally defined as a conditional probability model $p(\y|\x;\theta)$ parameterized by deep neural networks $\theta$.
Given an input sequence $\x = (x_1, x_2,\cdots, x_m)$ , a neural autoregressive model \cite{bahdanau2014neural, vaswani2017attention} predicts the target sequence $\y = (y_1, y_2, \cdots, y_n)$ sequentially based on the conditional distribution, which decomposes $p(\y|\x;\theta)$ by the autoregressive factorization:
\begin{equation}
    p_{\textit{AR}}(\y|\x;\theta)= \prod_{t=1}^{n} p(y_{t}|\bm{y}_{<t},\bm{x}; \theta), \nonumber
    \label{eqn:at}
\end{equation}
where $\theta$ is the set of model parameters.
Although such factorization achieved great success, its sequential prediction may cause high decoding latency and error accumulation during inference, especially for long sentences.

\paragraph{Non-autoregressive Translation.}
To solve above problems, \citet{gu2017non} proposed non-autoregressive Transformer based on conditional independence assumption among target tokens, which models $p(\y|\x;\theta)$ in a per-token factorization:
\begin{equation}
    p_{\textit{NAR}}(\y|\x;\theta)= \prod_{t=1}^{n}p(y_{t}|\bm{x}; \theta). \nonumber
    \label{eqn:nat}
\end{equation}
As a result, NAR models can boost up inference by predicting target words simultaneously, thereby improving the efficiency significantly.

However, as noted in \citet{gu2017non}, the target-side conditional independence assumption prohibits NAR models from capturing complex dependencies among target tokens, thereby significantly hurting accuracy. 
To mitigate this, a line of work proposes to modify the training objective~\cite{libovicky2018end, wang2019non, shao2020minimizing, ghazvininejad2020aligned, qian2020glancing, du2021order}, while other work uses latent variable to enhance modeling~\cite{kaiser2018fast, shu2020latent, bao2021non, bao2022latent}. 
Besides, several research proposes iterative-based models, which perform iterative refinement of translations based on previous predictions~\cite{lee2018deterministic, ghazvininejad2019mask, gu2019levenshtein, kasai2020non}. 
The most related work to this paper is \citet{hao2020multi}, which shows that utilizing an additional AR decoder could help the encoder of NAR models contain more linguistic knowledge.

%% file: 03methodology.tex
\section{Methodology}

In this section, we will dive deep into our simple yet effective multi-task learning framework, including model architecture and training scheme.

\paragraph{Model Architecture.}
The overall illustration of our approach is depicted in Figure~\ref{mtl-multi}. 
Specifically, for every NAR decoder layer, we introduce an auxiliary weak AR decoder, where each AR decoder is parameterized by one Transformer layer, being as weak as possible. 
In this case, these AR decoders will no longer capture the underlying structure of target sequences on their own, unless their NAR decoder layers can provide useful neural representations.
As a result, the NAR decoder layers can additionally learn from such informative task signals and become stronger to support the weak AR partners, being forced to capture sufficient context and dependency information.

\paragraph{Training Objective.}
Our training objective composes two parts for the NAR model of interest and the auxiliary weak AR decoders, respectively.
For the NAR part, we keep the original model-specific training objective unaltered. 
For instance, we apply CTC loss for CTC-based NAR models~\citep{saharia2020non}.
As for the AR decoders, we apply the cross-entropy loss for training.
The final loss is a weighted sum of the two components:
\begin{equation}
    \mathcal{L} = \lambda \mathcal{L}_{\textit{NAR}} + (1-\lambda)\sum_i^{N} \mathcal{L}^{(i)}_{\textit{AR}}, \nonumber
    \label{lossfunction}
\end{equation}
where $N$ is the number of NAR decoder layers, and $\mathcal{L}_{\textit{NAR}}$ and $\mathcal{L}_{\textit{AR}}$ represent the NAR loss and AR loss, respectively. The $\lambda$ is a predefined weight.

\begin{table*}[t]
    \centering
    \resizebox{0.8\linewidth}{!}{%
    \begin{tabular}{lcccccc}
    \toprule
    \multirow{2}{*}{Model} & \multicolumn{2}{c}{WMT14} & \multicolumn{2}{c}{WMT16} & IWSLT14 \\
           & \textsc{En-De}  & \textsc{De-En}  & \textsc{En-Ro}  & \textsc{Ro-En}   & \textsc{De-En} \\
    \midrule
    Vanilla-NAR \smallcitep{gu2017non}           & 17.69  & 21.47  & 27.29  & 29.06   & /   \\
    CMLM$_1$ \smallcitep{ghazvininejad2019mask}  & 18.05  & 21.83  & 27.32  & 28.20   & / \\
    Flowseq \smallcitep{flowseq}                 & 23.72  & 28.39  & 29.73  & 30.72   & 27.55 \\
    NAR-DCRF \smallcitep{nat_crf}                & 23.44  & 27.22  & /      & /       & 27.44 \\
    CTC \smallcitep{saharia2020non}              & 25.7   & 28.1   & 32.2   & 31.6    & /     \\
    AXE \smallcitep{ghazvininejad2020aligned}    & 23.5   & 27.9   & 30.75  & 31.54   & /   \\
    O$_\mathrm{A}$XE \smallcitep{du2021order}    & 26.1   & 30.2   & 32.4   & 33.3    & /   \\
    CNAT \smallcitep{bao2021non}                 & 25.56  & 29.36  & /      & /       & 31.15     \\
    GLAT \smallcitep{qian2020glancing}           & 25.21  & 29.84  & 31.19  & 32.04   & /   \\
    GLAT+CTC \smallcitep{qian2020glancing}       & 26.39  & 29.54  & 32.79  & 33.84   & /   \\
    DSLP \smallcitep{huang2022non}               & 27.02  & 31.61  & 34.17  & 34.60   & /     \\
    CMLM$_{10}$~\smallcitep{ghazvininejad2019mask} & 27.03 & 30.53 & 33.08  & 33.08   & / \\
    CMLM$_{10}$+MTL \smallcitep{hao2020multi} & \textbf{27.98} & 31.27 & 33.80 & 33.60  & / \\
    \midrule
    Transformer~(ours)      & 27.42   & 31.45  & 34.11  & 34.14   & 35.20 \\
    CTC~(ours)              & 26.27   & 29.60  & 32.63  & 33.47   & 33.91 \\
    CTC+MTL~(ours)          & 26.47   & 30.09  & 33.35  & 33.90   & 34.45 \\
    CTC+Our method          & 26.80   & 30.36  & 33.63  & 34.14   & 35.13 \\
    CTC+Our method \& Glancing Training & 27.25 & 30.70 & 33.88 & 34.73 & 35.15 \\
    \qquad beam search=20  & 27.75  & \textbf{31.81}  & \textbf{34.38} & \textbf{35.28}  &  \textbf{36.05}   \\
    \bottomrule
    \end{tabular}}
    \caption{Results of NAR models trained with knowledge distillation on test set of WMT14, WMT16 and IWSLT14. CMLM$_k$ refers to $k$ iterations of decoding. }
    \label{tab:main_result}
\end{table*}

\paragraph{Glancing Training.}
According to previous studies, glancing training~\cite{qian2020glancing} can considerably improve the translation quality of non-iterative NAR models. 
We apply glancing training technique to our method. 
More specifically, we first randomly sample reference tokens as NAR decoder inputs like~\citet{qian2020glancing}, 
and then let the weak AR decoder make predictions based on the NAR decoder hidden states.

\paragraph{Minimizing Training Cost.}
The major challenge of our method is additional training computational and memory overhead. 
To this end, we employ two techniques to reduce training costs:

\textit{(a) Parameter-sharing of AR decoders}.
    As all AR decoders are homogeneous, we can tie their parameters to reduce the total number of parameters.
    
\textit{(b) Layer dropout for AR decoders}.
    Simultaneously enabling every NAR decoder layer to pair its AR decoder partner is fairly inefficient.
    To this end, we randomly select half of the AR decoders, instead of all, for multi-task learning.
    
Both strategies help make the training cost affordable without losing accuracy gains.

\paragraph{Inference.}
We only use the NAR decoder for inference without any AR decoders.
The AR decoder is only used for training.
Therefore, our approach has no additional decoding overhead.

\begin{table}[t]
    \centering
    \resizebox{\linewidth}{!}{%
    \begin{tabular}{lccccc}
    \toprule
    \multirow{2}{*}{Model} & \multicolumn{2}{c}{WMT14} & \multicolumn{2}{c}{WMT16} &IWSLT14 \\
    & \textsc{En-De}  & \textsc{De-En} & \textsc{En-Ro}  & \textsc{Ro-En} & \textsc{De-En} \\
    \midrule
    Vanilla-NAR      & 17.79   & 22.02  & 27.84  & 29.35  &  28.32   \\
    \,\,\,\,\,\, + Our method    & \textbf{21.43}  & \textbf{25.85} & \textbf{29.88} & \textbf{30.89}      &  \textbf{32.26}  \\
    \midrule
    CTC     & 26.27     & 29.60    & 32.63  & 33.47  & 33.91 \\
    \,\,\,\,\,\, + Our method     & \textbf{26.80}  & \textbf{30.36}  & \textbf{33.63}  & \textbf{34.14} & \textbf{35.13} \\
    \bottomrule
    \end{tabular}
    }
    \caption{Results of applying our method to different NAR models, showing the generality of our method.}
    \label{tab:generality}
\end{table}

%% file: 04experiments.tex
\section{Experiments}

\paragraph{Experimental Settings.}
We conduct experiments on the most widely used machine translation benchmarks: WMT14 English-German (WMT14 \textsc{En-De}, 4.5M translation pairs), WMT16 English-Romanian (WMT16 \textsc{En-Ro}, 610K translation pairs) and IWSLT14 German-English~(IWSLT14 \textsc{De-En}, 160K translation pairs). We follow~\citet{gu2020fully} for data preprocessing and use BLEU~\cite{papineni2002bleu} as the evaluation metric.
To alleviate training difficulties, we use sequence-level knowledge distillation \cite{hinton2015distilling} for all datasets to alleviate multi-modality problem as in~\citet{gu2017non}.

\subsection{Main Results}

\paragraph{Our approach achieves superior results compared to existing strong NAR systems.}
Table~\ref{tab:main_result} presents our main results on the benchmarks. 
As seen, our method significantly improves the translation quality and outperforms other strong baseline models. 
Besides, when applying the glancing training technique, our method can result in further advancements.
Compared with CMLM, which employs iterative decoding, our model can achieve higher performance, while using single-step generation. 
\citet{hao2020multi}'s work is related to ours, which also utilizes a multi-task framework.
We reproduce their method on the CTC-based NAR model, and results show that our method can achieve greater improvements. 
Compared with the strong autoregressive teacher Transformer \cite{vaswani2017attention}, our model can further close the performance gap. And when decoding using beam search, our method can outperform Transformer on each dataset.

\paragraph{Our model-agnostic approach can help boost several classes of NAR models.}
We use Vanilla-NAR~\cite{gu2017non} and CTC~\cite{saharia2020non} models as baselines and apply our multi-task learning approach to each baseline model. The result is shown in Table \ref{tab:generality}. It can be seen that our method consistently and significantly improves the translation quality for each baseline model and each language pair. 
This illustrates the generality of our method.

\subsection{Analysis}

\begin{figure}[t]
  \centering
  \small
  \vspace{-10pt}
  \resizebox{0.75\linewidth}{!}{%
  \includegraphics[width=0.7\textwidth]{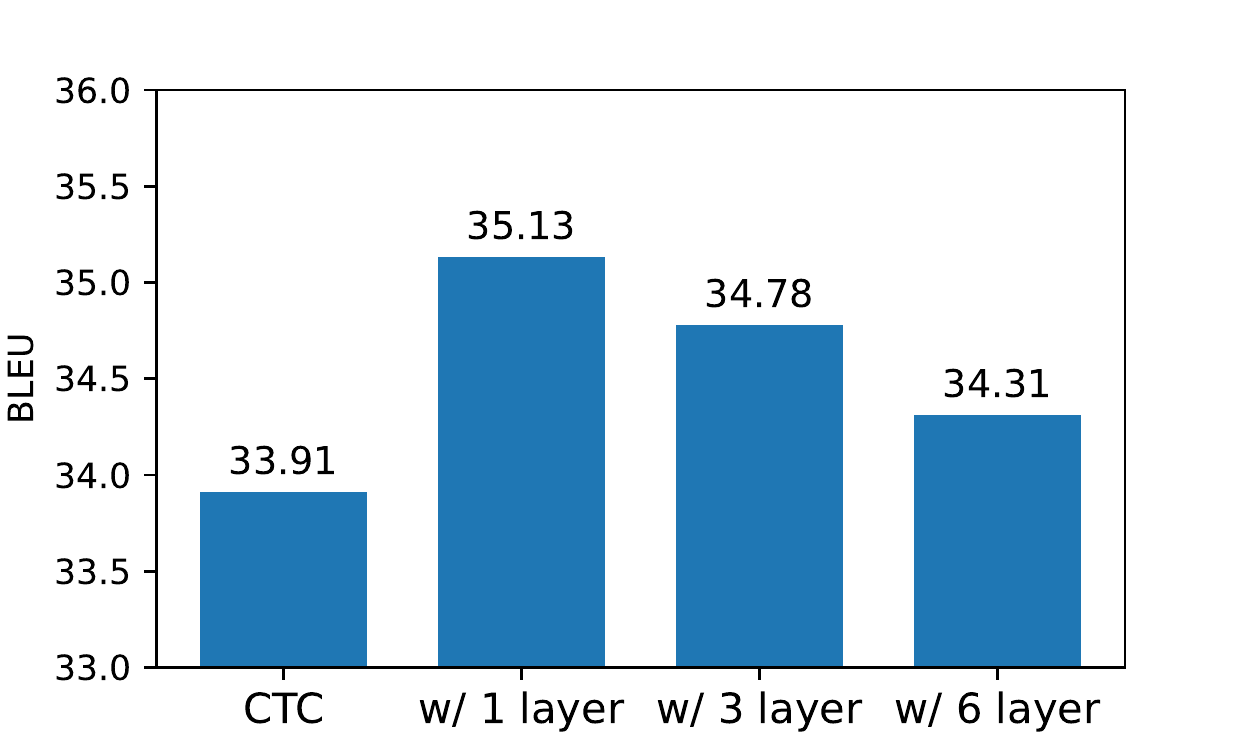} 
  }
  \caption{Results on the test of IWSLT14 to analyze the effectiveness of the number of AR decoder layers. We use CTC-based model as baseline, and w/ 1 layer means the AR decoder has 1 layer.}
  \label{tab:ablation_study}
  \vspace{-10pt}
\end{figure}

\paragraph{Does AR decoders being weak really matter?}
Recall that we let AR decoder be sufficiently weak to force NAR decoder to be strong. 
But how does the capacity of AR decoders affect the efficacy of our approach? 
We hence conduct experiments with the different number of AR decoder layers. e.g., 1, 3, and 6.
As demonstrated in Figure~\ref{tab:ablation_study}, each depth AR decoder can bring improvement, but as the number of AR decoder layers increases, the improvement effect for NAR gradually weakens. 
This verifies our motivation that a weaker AR decoder force NAR decoder to contain more useful information, in turn helping the NAR model.

\paragraph{Ablation study on the training cost optimization.}
We evaluate the impact of the proposed training cost reduction strategies.
As shown in Table~\ref{tab:reduce-trick}, after using the above two techniques, the number of parameters (83.8M vs 55.3M) and training time (31.2h vs 19.4h) are greatly reduced while keeping the model performance almost unchanged.

\begin{table}[t]
\centering
\tabcolsep 4pt
\small
\resizebox{\linewidth}{!}{%
\begin{tabular}{cccccc}
\toprule
 & \multirow{2}{*}{\shortstack{Param.\\Sharing}} & \multirow{2}{*}{\shortstack{Layer\\Dropout}} & \multirow{2}{*}{training time} & \multirow{2}{*}{\#params} & \multirow{2}{*}{BLEU}   \\ 
 \\
\midrule
\multirowcell{4}{\rotatebox[origin=c]{90}{Ours}} 
&                       &                   &    31.2h     &    83.8M           &  35.15     \\
& \hit                      &                   &   31.0h      &    55.3M       &  35.10     \\
&                       &   \hit                &   20.3h      &    83.8M       &  35.07     \\
& \hit                      & \hit              &   \textbf{19.4h}  &   \textbf{55.3M}    &  35.13     \\
\midrule
\multicolumn{3}{c}{CTC~\cite{saharia2020non}} &  17.3h    & 50.6M & 33.91 \\
\bottomrule
\end{tabular}
}
\caption{Study on training cost reduction.}
\label{tab:reduce-trick}
\end{table}


\begin{table}[t]
\small
\centering
\resizebox{\linewidth}{!}{%
\begin{tabular}{lcccc}
\toprule
Methods                 & (0,20]    & (20,40]   & (40,60]   & >60   \\
\midrule
Transformer             & 25.58     & 28.12     & 27.58     & 23.42  \\
CTC + Our method        & 24.77     & 27.54     & 27.18     & 25.07   \\      
\midrule
Gap                     & -0.81     & -0.58     & -0.40    & \textbf{+1.65} \\
\bottomrule
\end{tabular}
}
\caption{Results on the test of WMT14 \textsc{En-De} to analyze the performance differences of various target sentence length intervals. }
\label{tab:src_length}
\vspace{-8pt}
\end{table}

\begin{table}[t]
\centering
\resizebox{0.9\linewidth}{!}{%
\begin{tabular}{lcc}
\toprule
Methods                 & WMT14 \textsc{En-De}   & IWSLT14 \textsc{De-En}   \\
\midrule
Transformer             & 0.04\%        & 0.02\%    \\
\midrule       
Vanilla-NAR             & 16.2\%        & 6.94\%    \\
\,\,\,\,\,+Our method                 & \textbf{6.3\%}         & \textbf{2.90\%}    \\ 
\midrule 
CTC                     & 0.87\%        & 1.41\%    \\
\,\,\,\,\,+Our method                 & \textbf{0.11\%}        & \textbf{0.18\%}    \\
\bottomrule
\end{tabular}
}
\caption{Results of repeated token percentage. }
\label{tab:repeat_ratio}
\end{table}


\paragraph{Our approach helps handle lengthy sentences.}
To further analyze the performance differences on target sentences of different lengths, we divide the target sentences into buckets of different lengths. 
As shown in Table~\ref{tab:src_length}, as the sentence length increases, the performance gap between our model and the Transformer decreases.
Remarkably, our model outperforms Transformer when the target sentence length is greater than 60. 
Longer sentences mean that the model needs to deal with more complex contextual associations.
We conjecture that our proposed multi-task training method significantly improves the contextual information contained in the NAR hidden state, and thus has better performance on long sentence translation.

\paragraph{Our approach reduces token repetitions.}
We also study the rate of repeated tokens as in \cite{saharia2020non} to see to what extent our approach can tackle the multi-modality problem.
Table~\ref{tab:repeat_ratio} shows the repetition before and after applying our approach, demonstrating that our method consistently reduces the occurrence of repeated words by a significant margin.
Even when equipping CTC alone can alleviate the repetition issue, our approach can give rise to further improvements.

\begin{table}[t]
\centering
\resizebox{\linewidth}{!}{%
\begin{tabular}{lccc}
\toprule
\multirow{2}{*}{Model} & \multicolumn{2}{c}{WMT14}  &IWSLT14 \\
& \textsc{En-De}  & \textsc{De-En}  & \textsc{De-En} \\
\midrule
Transformer & 27.42 & 31.45 & 35.20 \\
Vanilla-NAR      & 11.02   & 15.13  & 17.72   \\
CTC     & 18.34     & 23.58    & 26.77 \\
\,\,+ Our method \& GLAT     & \textbf{24.14}  & \textbf{28.71}  & \textbf{31.48} \\
\bottomrule
\end{tabular}
}
\caption{Results without knowledge distillation. ``GLAT'' denotes glancing training.}
\label{tab:iwslt14-raw}
\end{table}

\paragraph{Performance without knowledge distillation.}
Despite knowledge distillation as a commonly-used workaround, it bounds the performance of NAR models under their AR teacher, along with the extra need to build teacher models.
To validate the effectiveness of our method in the raw data scenario, we conduct experiments on the WMT14 and IWSLT14 datasets without knowledge distillation. 
As shown in Table~\ref{tab:iwslt14-raw}, the baseline CTC model can be significantly enhanced by our approach, further closing the performance gap with the AR model.

\paragraph{Advantages of our method over other multi-task framework.}
\citet{hao2020multi}'s work also utilizes a multi-task framework, and our method can make greater improvements. We attribute this to the location and capacity of our multi-task learning module, i.e. the weak AR decoder.
For the location of the AR decoder, we argue that the decoder governs the generation, so placing the AR decoder upon the NAR decoder is supposed to more directly and explicitly improve the generation of NAR, while \citet{hao2020multi} is based on the NAR encoder output.
For the capacity of the AR decoder, we contend that the AR decoders should be as weak as possible, such that they can no longer model the target sequence on their own unless their NAR decoder layers can provide useful neural representations. In contrast, \citet{hao2020multi} do not elaborate on parameterization capacity and use a standard AR decoder. 

%% file: 05conclusion.tex
\section{Conclusion}

In this paper, we propose a multi-task learning framework for NAR. Along with the training of the weak AR decoder, the NAR hidden state will contain more contextual information, resulting in performance improvement. Experiments on WMT and IWSLT benchmarks show that our method can significantly and consistently improve the translation quality. When using beam search decoding, our CTC-based variant outperforms strong Transformer on all of the benchmarks, while introducing no additional decoding overhead.

%% file: 06limitation.tex
\section*{Limitations}

Our research's potential drawback is that it adds to the training burden. To tackle this problem, we introduce two techniques to reduce training costs. We greatly minimize the number of parameters that should be trained as well as the training time without sacrificing performance. Notably, our method does not introduce additional overhead for inference. Therefore, we can achieve a large performance improvement while maintaining the original fast decoding speed.

%% file: 07appendix.tex
\appendix

\section*{Appendix}
\label{sec:appendix}


\section{Training Hyperparameters}
We follow the normal hyperparameters used in NAR works. We design our NAR model with the base setting hyperparameters of Transformer \cite{vaswani2017attention}: both the encoder and the decoder has 6 layers, each layers has 8 attention head, and hidden dimension is 512. 
For the WMT tasks, we train the models with a batch size of 64K tokens and 300K updates. 
In the case of IWSLT tasks, we use a smaller batch size of 16K tokens, and set the maximum updates to 250K. For regularization, we set the dropout rate to 0.1 for WMT tasks and 0.3 for IWSLT tasks. We use Adam optimizer \cite{kingma2014adam} with $\beta=(0.9, 0.999)$. We employ weight decay of 0.01 and label smoothing of 0.1. For the shallow AR decoder, we set the number of decoder layers to 1. We set the hyperparameter $\lambda$ used in Eq.~\ref{lossfunction} to 0.5. To obtain robust results, we averaged the last 5 best checkpoints, following~\citet{vaswani2017attention}. All models are implemented on \texttt{fairseq}~\cite{ott2019fairseq}.